\documentclass{article}

\usepackage[final]{neurips_2022}

\usepackage[utf8]{inputenc} 
\usepackage[T1]{fontenc}    
\usepackage{hyperref}       
\usepackage{url}            
\usepackage{booktabs}       
\usepackage{amsfonts}       
\usepackage{nicefrac}       
\usepackage{microtype}      
\usepackage{xcolor}         

\usepackage{amsthm}
\usepackage{amsmath}
\usepackage{bbm}
\theoremstyle{definition}
\newtheorem{definition}{Definition}[section]
\newtheorem{theorem}{Theorem}[section]

\usepackage{xspace}
\usepackage{graphbox}
\usepackage[pro]{fontawesome5}
\usepackage{wrapfig}
\usepackage{titletoc}
\usepackage[page,header]{appendix}
\usepackage[inline]{enumitem}
\usepackage{subfig}
\usepackage[ruled,vlined,linesnumbered]{algorithm2e}
\usepackage{tikz}
\usetikzlibrary{arrows, automata, positioning}

\usepackage[scaled=0.77]{beramono}
\usepackage[T1]{fontenc}

\definecolor{gold}{RGB}{255,215,0}
\definecolor{goldenrod}{RGB}{218,165,32}
\definecolor{maroon}{rgb}{0.5, 0.0, 0.0}

\newcommand{\tuple}[1]{\langle #1 \rangle}

\newcommand{\althide}[1]{}

\newcommand{\removehide}[1]{}

\newif\ifcomments
\commentstrue

\ifcomments

\newcommand{\commentsm}[1]{\textcolor{purple}{({\bf SM:} #1)}}
\newcommand{\commentsmhide}[1]{}

\newcommand{\commenttk}[1]{\textcolor{teal}{({\bf TK:} #1)}}
\newcommand{\commenttkhide}[1]{}

\newcommand{\commental}[1]{\textcolor{orange}{({\bf AL:} #1)}}
\newcommand{\commentalhide}[1]{}

\newcommand{\commentzc}[1]{\textcolor{gray}{({\bf ZC:} #1)}}
\newcommand{\commentzchide}[1]{}

\newcommand{\commentpv}[1]{\textcolor{maroon}{({\bf PV:} #1)}}
\newcommand{\commentpvhide}[1]{}

\else

\newcommand{\commentsm}[1]{}
\newcommand{\commentsmhide}[1]{}

\newcommand{\commenttk}[1]{}
\newcommand{\commenttkhide}[1]{}

\newcommand{\commental}[1]{}
\newcommand{\commentalhide}[1]{}

\newcommand{\commentzc}[1]{}
\newcommand{\commentzchide}[1]{}

\newcommand{\commentpv}[1]{}
\newcommand{\commentpvhide}[1]{}
\fi

\newcommand{\pr}[0]{\mathrm{Pr}}
\newcommand{\ttt}[1]{\texttt{#1}}
\def\stone(#1)[#2][#3][#4]{
    \begin{scope}[shift={(#1)},rotate=#2,scale=#3, transform shape]
        \draw[myshade,opacity=#4] (15:3pt) -- (45:4pt) -- (89:5pt)--(135:6pt) -- (150:3pt) -- (182:4pt)--(235:6pt) -- (280:3pt) -- (330:2pt) -- cycle;
    \end{scope}
}

\def\mine(#1)[#2]{
    \begin{scope}[shift={(#1)},transform shape]
        \stone(7pt,8pt)[55][1.1][#2]
        \stone(0pt,0pt)[45][0.7][#2]
        \stone(14pt,14pt)[60][0.7][#2]
        \stone(0pt,15pt)[110][0.5][#2]
        \stone(15pt,2pt)[-20][0.7][#2]
        \stone(7.5pt,0pt)[90][0.5][#2]
    \end{scope}
}

\newcommand{\depot}{\text{\faHome}\xspace}
\newcommand{\gold}{
    \begin{tikzpicture}[
        myshade/.style={
            goldenrod!50,
            draw,
            ultra thin,
            rounded corners=.5mm,
            top color=gold!20,
            bottom color=gold!20!goldenrod
        },
        interface/.style={
            postaction={draw,decorate,decoration={border,angle=-135,
            amplitude=0.3cm,segment length=2mm}}
            }
        ]
        \stone(0, 0)[80][0.9][1]
    \end{tikzpicture}
    \xspace
}

\title{Noisy Symbolic Abstractions for Deep RL: \\ A case study with Reward Machines}

\author{
    Andrew C. Li\thanks{Equal Contribution} \\
    University of Toronto \\
    Vector Institute \\
    \And
    Zizhao Chen\textsuperscript{$\ast$} \\
    University of Toronto \\
    Vector Institute \\
    \And
    Pashootan Vaezipoor \\
    University of Toronto \\
    Vector Institute \\
    \And
    Toryn Q. Klassen\thanks{Also affiliated with the Schwartz Reisman Institute for Technology and Society} \\
    University of Toronto \\
    Vector Institute \\
    \And
    Rodrigo Toro Icarte \\ 
    Pontificia Universidad Católica de Chile \\
    Centro Nacional de Inteligencia Artificial \\
    \And
    Sheila A. McIlraith\textsuperscript{$\dagger$} \\
    University of Toronto \\
    Vector Institute \\
}

\begin{document}

\maketitle

\begin{abstract}
Natural and formal languages provide an effective mechanism for humans to specify instructions and reward functions. 
We investigate how to generate policies via RL when reward functions are specified in a symbolic language captured by Reward Machines, an increasingly popular automaton-inspired structure.  We are interested in the case where the mapping of environment state to a symbolic (here, Reward Machine) vocabulary -- commonly known as the \emph{labelling function} -- is uncertain from the perspective of the agent. We formulate the problem of policy learning in Reward Machines with noisy symbolic abstractions as a special class of POMDP optimization problem, and investigate several methods to address the problem, building on existing and new techniques, 
the latter focused on predicting Reward Machine state, rather than on grounding of individual symbols. We analyze these methods and evaluate them experimentally under varying degrees of uncertainty in the correct interpretation of the symbolic vocabulary.
We verify the strength of our approach and the limitation of existing methods via an empirical investigation on both illustrative, toy domains and partially observable, deep RL domains.
\end{abstract}

\begin{section}{Introduction}
\label{sec:introduction}

Language is playing an increasingly important role in Reinforcement Learning (RL). 
It serves as a mechanism for expressing human-interpretable reward functions as well as for conveying language-based advice and instructions (e.g., \citep{luketina2019survey,tuli2022learning, huang2022language}).
An oft-cited challenge that faces the integration of language and machine learning is the \emph{symbol grounding problem} – the problem of associating symbols or words in a language (e.g. “cat” or “chair”) or properties expressed in a language (e.g, “the robot is holding the cup”) with states of the world – especially if those states are generated via a different modality such as vision, haptics or audio. Indeed the association of the symbols and sentences of language with observed state is contextual, complex and noisy, and it is this observation that broadly motivates the work in this paper.


Our general concern is with learning a policy using RL when language is used to specify reward-worthy behaviour or task instructions over a vocabulary – a set of symbols – that captures a purposeful abstraction of the observed environment state. We are specifically concerned with the case where the interpretation of this vocabulary is noisy or uncertain, and how to perform RL effectively in this context. This causes the RL agent to be uncertain about whether it is making progress on garnering rewards and/or following instructions that have been conveyed via language. For example, consider that an autonomous vehicle gets negative reward for driving through a red light, but where the determination of whether "red light" is true in the current state can be compromised by a slush-covered camera lens, specularities from the sun, or occlusion from a truck in front of the vehicle. 


For the purposes of this paper, we examine this problem in the context of Reward Machines (RMs), automata-like structures that provide a normal-form representation for (non-Markovian) reward functions, human preferences, and instructions \citep{DBLP:conf/icml/IcarteKVM18,icarte2022reward}. 
The virtue of RMs is that they allow for reward specification in a diversity of (formal) languages that have corresponding automata representations  \citep{camamacho2019ijcai}, following Chomsky's hierarchy (e.g., \cite{hopcroft-motwani-ullman2007}). Further, RMs are typically paired with learning algorithms that exploit the reward function structure exposed in the automata to significantly improve sample efficiency. RMs have historically realized symbol grounding via so-called \emph{labelling functions} that map environment states into a vocabulary of properties (e.g., “at-location-A”, “light-is-red”)  that are \emph{propositional} – either true or false in the state. RMs have proven effective when such functions faithfully reflect the intended interpretation of the propositional symbols, but when environments are partially observable or when labelling functions reflect uncertainty or sensors are noisy (as many real-world detectors are)
they may not be adequate to determine the truth or falsity of a particular state property with certainty. 

The use of formal languages in (deep) RL, including Linear Temporal Logic and Reward Machines, is a topic of broad interest because of the benefits accrued by exploiting the compositional syntax and semantics of formal languages, their connections with formal system verification, and correct-by-construction synthesis from formal specification \citep[e.g.,][]{aksaray2016q,li2017reinforcement,li2018policy, littman2017environment,DBLP:conf/icml/IcarteKVM18,tor-etal-neurips19, hasanbeig2018logically,deeplcrl,alur2019,DBLP:conf/ijcai/0005T19,kuo2020encoding,HasanbeigKA20Deep,xu2020joint,xu2020active,furelos2020inductionAAAI,rens2020learning,DBLP:conf/kr/GiacomoFIPR20,jiang2020temporal, neary2020reward, shah2020planning, velasquez2021learning,defazio2021learning,zheng2021lifelong,camacho2021reward,corazza2022reinforcement,liu2022skill}.




The contributions of this paper are as follows:

        \begin{itemize}[leftmargin=1em]
        \item 
        We study the problem of learning policies when (non-Markovian) reward functions are expressed in a symbolic language captured by a Reward Machine, but where the interpretation of the vocabulary of that language -- the symbols -- is noisy. We present a novel formulation of this problem as optimizing a POMDP, where the POMDP state comprises both the concrete observable environment state and the hidden Reward Machine state.

        \item We investigate a suite of deep RL methods for our proposed problem formulation and highlight their limitations under various conditions. 
        
        
        
        \item We propose a novel deep RL algorithm, \emph{Reward Machine State Modelling}, that makes robust decisions by being aware of the uncertainty in its interpretation of the vocabulary. 
        
        \item We verify the strength of our approach and limitations of existing methods via an empirical investigation using illustrative, toy examples and partially observable, deep RL image domains. 

        \end{itemize}

%




        
\end{section}
\begin{section}{Preliminaries}



The goal of reinforcement learning is to learn an optimal policy $\pi$ by interacting with an environment \citep{sutton2018reinforcement}. Typically, the environment is modelled as a \emph{Markov Decision Process} (MDP) \citep{bellman1957markovian}, a tuple $\langle {S},{T},{A},{P},{R},\gamma, \mu \rangle$. ${S}$ is the set of \emph{states}, ${T} \subseteq {S}$ is the set of \emph{terminal states}, ${A}$ is the set of \emph{actions}, ${P}(s' | s, a)$ is the \emph{transition probability distribution}, ${R}: {S} \times {A} \times {S} \rightarrow \mathbb{R}$ is the \emph{reward function}, $\gamma$ is the \emph{discount factor}, and $\mu$ is the \emph{initial state distribution}.

Reward machines are an automata-based representation of (non-Markovian) reward functions \citep{DBLP:conf/icml/IcarteKVM18,
icarte2022reward} that provide a normal-form representation for reward specification in a diversity of formal languages \citep{camamacho2019ijcai}. Given a \emph{vocabulary} -- a finite set of propositions $\mathcal{AP}$ representing abstract (possibly
human interpretable) properties or events in the environment, RMs specify temporally extended rewards
over these propositions while exposing the compositional reward structure to the learning agent. Formally, Reward Machines, which we take to mean here \emph{simple Reward Machines} from \cite{icarte2022reward}, are defined as follows:

\begin{figure}[t]%
\centering
    \centering
    \parbox{0.49\textwidth}{
    \scalebox{0.65} {
    \begin{tikzpicture}[
    myshade/.style={
        goldenrod!50,
        draw,
        ultra thin,
        rounded corners=.5mm,
        top color=gold!20,
        bottom color=gold!20!goldenrod
    },
    interface/.style={
        postaction={draw,decorate,decoration={border,angle=-135,
        amplitude=0.3cm,segment length=2mm}}
        }
    ]

    \foreach \i in {0,...,4} {
        \draw [black] (\i,0 - 0.05) -- (\i,4 + 0.05);  
    }
    \foreach \i in {0,...,4} {
        \draw [black] (0,\i) -- (4,\i);
    }
    \draw [black, line width=0.1cm] (0, 0) rectangle (4, 4) {};

    \mine(3.25, 3.25)[1]
    \mine(3.25, 2.25)[1]
    \mine(3.25, 1.25)[1]
    \mine(3.25, 0.25)[1]
    
    \node[scale=1.5] at (0.5,3.5) {\faRobot};
    \node[scale=1.5] at (0.5,0.5) {\faHome};

\end{tikzpicture}

    }}%
    \hspace{-6em}
    \parbox{0.49\textwidth}{
    \scalebox{0.8} {
    \begin{tikzpicture}[
    ->, 
    >=stealth', 
    node distance=3cm, 
    every state/.style={thick, fill=gray!10}, 
    ]
    \node[state, initial] (u0) {}; 
    \node[state, right=2.5cm of u0] (u1) {}; 
    \node[state, right of=u1, accepting] (u2) {}; 
    \node[state, below right=1.5cm of u0, accepting] (u3) {}; 
    \draw (u0) edge[loop above] node{$\neg \; \gold \land \neg \;\depot$ / 0} (u0);
    \draw (u0) edge[above] node{$\gold \land \neg \; \depot$ / 0} (u1);
    \draw (u0) edge[below left] node{\depot / 0} (u3);
    \draw (u1) edge[loop above] node{$\neg \;\depot$ / 0} (u1);
    \draw (u1) edge[above] node{$\depot$ / 1} (u2);
    
\end{tikzpicture}
    
    }}
    \caption{
    %
    %
   The \emph{Mining} Domain. \textbf{Left:} A robot is tasked with digging up a stash of gold (found in any one of the four right-most squares) and then delivering it to the depot at the bottom left. Visiting the depot ends the episode. \textbf{Right:} A Reward Machine describing this task. The atomic propositions are $\mathcal{AP} = \{ \text{\protect\gold}, \text{\depot} \}$, where \protect\gold holds when the robot performs a "dig" action in a gold square and \depot holds when the robot is at the depot. A directed edge between RM states $u$ and $v$ labelled $\phi/r$ means that when $\phi$ (a propositional formula over $\mathcal{AP}$) is satisfied while the agent is in RM state $u$, the agent gets reward $r$ and transitions to $v$.}
\label{fig:mining}%
\vspace{-1em}
\end{figure}
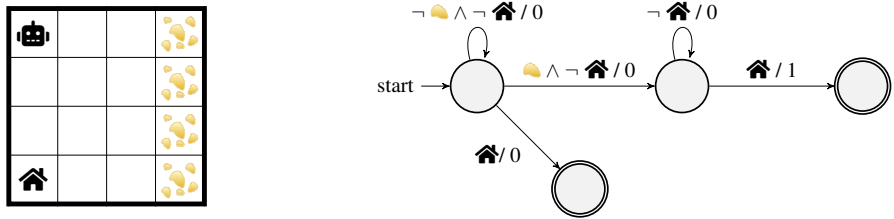

\begin{definition}[Reward Machine, RM] 
A simple \emph{Reward Machine} is a tuple 
$\langle U, u_0, F, \mathcal{AP}, \delta_t, \delta_r \rangle$, where $U$ is a finite set of states, $u_0\in U$ is the initial state, $F$ is a finite set of terminal states (disjoint from $U$), $\mathcal{AP}$ is a finite set of atomic propositions, $\delta_u : U \times 2^\mathcal{AP} \rightarrow (U \cup F)$ is the state-transition function, and $\delta_r : U \times 2^\mathcal{AP} \rightarrow \mathbb{R}$ is the state-reward function.

\end{definition}



\noindent{\bf Running Example:} The \emph{Mining} domain depicted in Figure~\ref{fig:mining} 
(adapted from \citep{corazza2022reinforcement}) illustrates the use of RMs. 
The agent's state is its current square on the grid, and its task is to obtain at least one ore of gold (found at a gold square), then deliver it to the depot. Visiting the depot ends the episode.
Figure~\ref{fig:mining} (right) shows a possible RM for this task over propositions $\mathcal{AP} = \{\gold, \depot\}$.


RMs can be used in place of a standard reward function in an MDP or POMDP $\mathcal{M}$ with the help of a \emph{labelling function} $\mathcal{L}:S\times A\times S\to 2^\mathcal{AP}$. The labelling function grounds abstract propositions from $\mathcal{AP}$ into the environment $\mathcal{M}$ by mapping transitions $(s_t, a_t, s_{t+1})$ into the subset of propositions that hold for that transition. In the \emph{Mining} example, a labelling function $\mathcal{L}$ evaluates \!\gold\!\! to true when the agent digs at a gold square and evaluates \depot to true when the agent is at the bottom-left square.

At each (non-final) timestep $t$, if the environment transition is $(s_t, a_t, s_{t+1})$ and the RM state is $u_t \in U$ (initially set to $u_0$), then the RM state is updated to $u_{t+1} = \delta_u(u_t, \mathcal{L}(s_t,a_t,s_{t+1}))$ and the agent receives reward $r_t=\delta_r(u_t, \mathcal{L}(s_t,a_t,s_{t+1}))$. 
This process of using an RM with an MDP is equivalent to using a larger MDP -- which we will call the \emph{Reward Machine MDP} -- that has a standard (Markovian) reward function but with state space $S\times U$ \citep[see][Observation 1]{icarte2022reward}.

\begin{definition}[{Reward Machine MDP, RM-MDP}] 
\label{def:rmmdp}
Given an MDP without a reward function $\tuple{S,T,A,P,\gamma,\mu}$, an RM $\tuple{ U, u_0, F, \mathcal{AP}, \delta_t, \delta_r}$, and a labelling function $\mathcal{L}:S\times A\times S\to 2^\mathcal{AP}$, the corresponding \emph{Reward Machine MDP} is an MDP $
\tuple{S',T',A, P',R',\gamma, \mu'}$ where $S' = S \times U$ is the state space, $T' = (T\times U)\cup (S\times F)$ are terminal states, $A$ is the action space, $P'((s_{t+1}, u_{t+1}) | (s_t,u_t),a_t) = P(s_{t+1}|s_t,a_t) \cdot \mathbbm{1}[\delta_u(u_t, \mathcal{L}(s_t, a_t, s_{t+1})) = u_{t+1}]$ are the transition probabilities, $R'((s_t,u_t),a_t,(s_{t+1},u_{t+1})) = \delta_r(u_t, \mathcal{L}(s_t, a_t, s_{t+1}))$ is the reward function, $\gamma$ is the discount factor, and $\mu'(s,u) = \mu(s)\mathbbm{1}[u = u_0]$ is the initial state distribution. 
\end{definition}




RMs are commonly integrated into deep RL algorithms by using the labelling function $\mathcal{L}$ to compute the RM state $u_t$, which is provided as input to the policy along with $s_t$. In the \emph{Mining} example, the RM state $u_t$ acts as salient memory, capturing whether the agent has previously obtained gold.

\end{section}

\begin{section}{Decision-making under Uncertain Propositions}
    



The majority of existing RL methods that employ structured 
reward function specifications, such as RMs, 
assume \emph{perfect} labelling functions in their decision-making. By exploiting the ground-truth propositional values, the true state of the RM is always available to the policy. 
Unfortunately, faithful evaluation of these abstracted propositions is unrealistic in many real-world settings; factors such as sensor noise, modelling errors, partial observability or insufficient data may all lead to incorrect 
evaluation of these propositions. Critically, the agent's estimate of the RM state may rapidly diverge from reality as a result of such errors, potentially provoking unintended behaviours. 


\noindent{\bf Example continued:} Consider again the \emph{Mining} example in Figure~\ref{fig:mining} when the agent is unable to consistently determine if an ore is gold or another less valuable mineral in the absence of a perfect labelling function. Now, the agent cannot be certain that it has obtained gold, and a false positive belief may cause the agent to head to the depot, failing the task. A robust policy should be cognizant of this uncertainty, perhaps digging at many squares before visiting the depot to maximize its probability of successfully completing the task. Unfortunately, efficient realization of this behaviour is inherently non-Markovian, as it requires the agent to remember which squares were previously mined.




Our goal is to investigate how agents can learn these robust behaviours without a perfect labelling function. We formulate this problem in RM-MDPs as solving a special type of POMDP.


\begin{subsection}{Problem Formulation}






Given an MDP without a reward function $\tuple{S,T,A,P,\gamma,\mu}$ and an RM $\tuple{ U, u_0, F, \mathcal{AP}, \delta_t, \delta_r}$, and assuming the existence of a (hidden) labelling function $\mathcal{L}:S\times A\times S\to 2^\mathcal{AP}$,
our objective is to learn a policy that maximizes the expected discounted return according to the corresponding RM-MDP (Definition~\ref{def:rmmdp}) without having access to $\mathcal{L}$. 


We formulate this goal as optimizing a POMDP with state space $S \times U$, and transitions matching those in the RM-MDP. However, while an RM-MDP treats RM states $u_t \in U$ as an observable component of the state, only $s_t \in S$ is observable but not $u_t$, because $u_t$ depends on the ground truth labelling function. We call such a POMDP an \emph{Uncertain-Proposition Reward Machine MDP}.

\begin{definition}[Uncertain-Proposition Reward Machine MDP, URM-MDP] Given an MDP without a reward function $\tuple{S,T,A,P,\gamma,\mu}$, an RM $\tuple{ U, u_0, F, \mathcal{AP}, \delta_t, \delta_r}$, and a labelling function $\mathcal{L}:S\times A\times S\to 2^\mathcal{AP}$, the corresponding \emph{URM-MDP} is a  POMDP $\langle S', \Omega, T', A, P', O, R', \gamma, \mu' \rangle$, where $S'$, $T'$, $A$, $P'$, $\gamma$ and $\mu'$ are as in the corresponding RM-MDP (Definition~\ref{def:rmmdp}), $\Omega = S $ is the observation space, and $O(s|(s_t, u_t), a_t) = \mathbbm{1}[s=s_t]$ are the observation probabilities.
\end{definition}

We similarly define an \emph{Uncertain-Proposition Reward Machine POMDP} (URM-POMDP with Definition~\ref{urmpomdp} in the Appendix), where the underlying environment is only partially observable, but the labelling function $\mathcal{L}$ remains a function of $S \times A \times S$. 


\medskip


\end{subsection}

\end{section}

\section{Analysis of Prior Approaches}
\label{sec:methods}

A number of 
prior approaches 
have considered uncertainty in the values of propositions, though mainly for tabular MDPs. Unfortunately, their effectiveness remains unclear when applied to complex deep RL settings where the number of states is infinite, high-level events are challenging to perfectly model, or where the environment is only partially observable. 




One type of approach models the true labelling function $\mathcal{L}$ with an approximate labelling function, $\hat{\mathcal{L}}: S\times A\times S \times \mathcal{AP} \rightarrow [0,1]$ in URM-MDPs, and $\hat{\mathcal{L}}: H\times A\times \Omega \times \mathcal{AP} \rightarrow [0,1]$ in URM-POMDPs,
 that assigns probabilities to each proposition. Such an approximate labelling function might be obtained from external sensors, handcrafted heuristics, offline data, or be trained online together with the policy. Then, the predicted propositions are used to generate an estimate (or a belief) of the current RM state using one of the following methods. One appeal of this technique is that the approximate labelling functions can in principle be transferred to new environments and RMs, so long as the high-level propositions remain the same.  


\textbf{\ttt{Thresholding:}} {Inspired by, e.g., \cite{da2019active, ghasemi2020task}, and \cite{tuli2022learning}, 
this method discretizes every proposition $p_i$ at each timestep $t$ to its most likely value under $\hat{\mathcal{L}}$. In URM-MDPs, $\hat{p}_{i,t} = \mathbbm{1}[\hat{\mathcal{L}}(s_t,a_t,s_{t+1}, p_i) \ge 0.5]$ and the $\hat{p}_{i,t}$ values are then used to produce a discrete prediction of an RM state $\hat{u}_t$. The learned policy takes the form $\pi(a_t | s_t, \hat{u}_t)$. In URM-POMDPs, we simply change the approximate labelling function to the correct form $\hat{\mathcal{L}}: H\times A\times \Omega \times \mathcal{AP} \rightarrow [0,1]$ and change the policy to condition on histories $h_t$ rather than states $s_t$. 

\textbf{\ttt{Independent Belief Updating:}} Inspired by, e.g., \cite{ding2011ltl} and \cite{verginis2022joint}, this approach maintains a probabilistic belief $\tilde{u}_t$ over the current RM state under the simplifying assumption that predictions of propositions $p_{i,t}$ are independent of one another and across all timesteps. In URM-MDPs, 
$\tilde{u}_t$ is updated
each timestep for all $u \in U$, given an experience $(s_t, a_t, s_{t+1})$, via
\[
\tilde{u}_{t+1}(u) = \sum_{\sigma_t \in 2^{\mathcal{AP}}, u_t \in U}{\pr(\sigma_t | s_t, a_t, s_{t+1})\tilde{u}_t(u_t) \mathbbm{1}[\delta_u(u_t, \sigma_t) = u]} \]
where $\pr(\sigma_t | s_t, a_t, s_{t+1})$ is the probability of a particular propositional assignment $\sigma_t$ under $\hat{\mathcal{L}}$. The learned policy then takes the form $\pi(a_t | s_t, \tilde{u}_t)$. In URM-POMDPs, we change $\hat{\mathcal{L}}$ to the correct form and change the policy to condition on histories $h_t$ rather than states $s_t$.



\textbf{\ttt{Recurrent Policy:}} Following \cite{kuo2020encoding}, we also consider a third approach that directly solves the URM-MDP or URM-POMDP as a general partially observable RL task 
by learning
a recurrent policy of the form $\pi(a_t | s_0, a_0, ..., s_{t-1}, a_{t-1}, s_t)$ in URM-MDPs and $\pi(a_t | o_0, a_0, ..., o_{t-1}, a_{t-1}, o_t)$ in URM-POMDPs. Notice that the policy does not condition on the RM state (or an estimate of it); the non-Markovian task structure is instead handled by conditioning on the entire history. Thus, propositional values (either ground-truth, or estimated) are not required for this approach.

\subsection{Pitfalls in URM-MDPs}
\label{sec:remarks-urm-mdp}
In URM-MDPs, an approximate labelling function $\hat{\mathcal{L}}: S \times A\times S\times  \mathcal{AP} \rightarrow [0,1]$ can in principle perfectly model the ground-truth labelling function $\mathcal{L}$. In this ideal case, both \ttt{Thresholding} and \ttt{Independent Belief Updating} correctly estimate the true RM state, inheriting any optimality guarantees of the RL algorithm used to train the policy \citep{icarte2022reward}.

\begin{wrapfigure}{R}{0.5\textwidth}
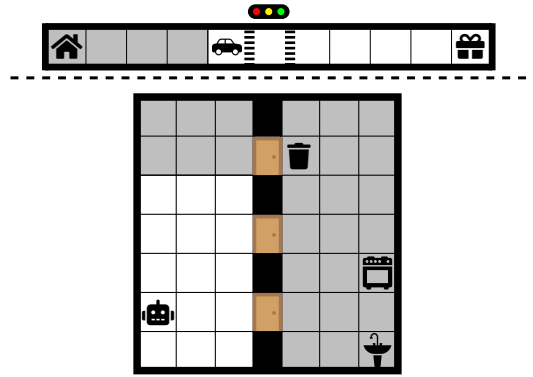

    \centering
    \begin{minipage}{0.5\textwidth}
        \centering
        \scalebox{0.9}{
        \input{figs/traffic}
        }
    \end{minipage}
    \\
    \begin{tikzpicture}
    \draw [dashed, line width=0.04cm] (-3,0) -- (4,0);
    \end{tikzpicture}
    \\
    \begin{minipage}{0.5\textwidth}
        \centering
        \input{figs/kitchen}
    \end{minipage}

    \caption{Two \emph{partially observable} RM environments that challenge existing methods. Shaded squares are unobservable to the agent. \textbf{Top:} The \emph{Traffic Light} domain. The agent must drive through an intersection to collect a package, and then drive through again to return home. However, the agent can only see forwards and must avoid crossing a red light. \textbf{Bottom:} The \emph{Kitchen} domain. It's the end of the day and a household robot must go to the kitchen to clean up by ensuring dishes are washed, the table is cleared, and the garbage is empty. Some chores might not require the robot's attention, e.g. the residents may have not used any dishes. This occurs with a 1/3 chance for each chore, and that chore is considered done from the start. However, the agent cannot observe what chores remain to be done until entering the kitchen.} 
    \label{fig:po_envs}%
    \vspace{-3em}
\end{wrapfigure}

Unfortunately, perfectly modelling the labelling function can be impractical in complex, real-world settings. 
\ttt{Thresholding} is resistant to small errors in $\hat{\mathcal{L}}$, but only estimates a single, discrete RM state and cannot represent the uncertainty in its prediction.
\ttt{Independent Belief Updating} aims to address this by estimating a belief over RM states, but the independence assumption often fails to hold 
for noisy predictions under $\hat{\mathcal{L}}$. In the \emph{Mining} example, repeatedly digging in a square that is erroneously believed to contain gold will increase the perceived probability of having gold under \ttt{Independent Belief Updating}. However, digging more than once in any one square is pointless since either all digs will produce gold, or none will. 

\ttt{Recurrent Policy} treats URM-MDPs as general POMDPs and avoids approximating the labelling function altogether. While it retains any convergence guarantees afforded by partially observable RL algorithms, history-based policies are notoriously hard and sample-inefficient to train in practice.


\subsection{Pitfalls in URM-POMDPs}
\label{sec:remarks-urm-pomdp}

Significant issues arise with \ttt{Thresholding} and \ttt{Independent Belief Updating} when the underlying environment is partially observable. First, even the Bayes-optimal approximate labelling function $\hat{\mathcal{L}^*}(h_t, a_t, o_t, p_i)$ is not necessarily identical to the ground-truth labelling function $\mathcal{L}$, since the propositions depend on unobservable states $s_t$. Second, the current RM state $u_t$ generally cannot be determined with certainty given the history $h_t$ due to the previous remark. Thus, \ttt{Thresholding} may predict an incorrect RM state, even with this ideal approximate labelling function $\hat{\mathcal{L}^*}$. 

The \emph{Traffic Light} example in Figure~\ref{fig:po_envs} highlights how a policy might abuse this, leading to unsafe behaviour. The agent must safely drive through an intersection with a traffic light, but cannot observe the light colour when traversing the intersection \emph{in reverse}. If the light tends to be green more often than red, \ttt{Thresholding} would predict that the agent did \emph{not} pass a red light. Hence, the agent would be incentivized to drive backwards through the intersection while ignoring the light colour, rather than waiting for a green light.

\ttt{Independent Belief Updating} may similarly fail to model a correct belief over RM states, even with the ideal $\hat{\mathcal{L}^*}$. We demonstrate this in the \emph{Kitchen} example in Figure~\ref{fig:po_envs}, where a cleaning agent cannot initially observe which chores require its attention from outside the kitchen. The Bayes-optimal $\hat{\mathcal{L}^*}$ assigns $1/3$ probability that each chore is done on each timestep while outside the kitchen. If the agent wanders outside the kitchen for $t$ timesteps, \ttt{Independent Belief Updating} would predict that the dishes were washed at some point with probability $1 - (2/3)^t$. The correct probability is $1/3$, since the propositions are linked over time -- if the dishes are unwashed now, they will remain unwashed on the next step unless the agent washes them. \ttt{Independent Belief Updating} ignores this dependence and eventually predicts that all chores are done with probability close to 1, despite the agent never entering the kitchen. Unfortunately, modelling the correct temporal dependencies between propositions is hard -- in general, one may need to model the full joint distribution over all propositions up to time $t$, i.e. $\{ p_{i, t'}: 1 \le i \le |\mathcal{AP}|, 0 \le t' \le t\}$, given the history $h_t$. 

Finally, we note that \ttt{Recurrent Policy} does not significantly distinguish between URM-POMDPs and URM-MDPs; a history-based policy is learned in each case. Thus, our remarks from the URM-MDP setting carry over to URM-POMDPs as well.

\section{Proposed Approach}
\label{sec:approach}
We propose a new approach, \emph{Reward Machine State Modelling} (\textbf{\ttt{RMSM}}) to address the shortcomings of prior approaches in URM-(PO)MDPs. Our key observation is 
that the utility of modelling the labelling function is to accurately predict the RM state.
Thus, we propose to directly learn the belief over RM states, given the history, using a recurrent neural network (RNN) $g_\phi$ trained via supervised learning. The policy $\pi_\theta$ then conditions on this predicted belief in place of the ground-truth RM state.

We describe \ttt{RMSM} in detail for URM-MDPs in Algorithm~\ref{algo:rmsm}. In this setting, the network inputs are $g_\phi(u_t | s_0, a_0, \ldots, s_{t-1}, a_{t-1}, s_t)$ for the RM belief RNN and $\pi_\theta(a_t | s_t, \tilde{u}_t)$ for the policy, where $\tilde{u}_t = g_\phi(\cdot | s_0, a_0, \ldots, s_{t-1},a_{t-1},s_t)$ is the predicted belief over RM states. Note that $\tilde{u}_t$ can be compactly represented with a vector of $|U|$ probabilities. In URM-POMDPs, the policy $\pi_\theta$ instead conditions on $(h_t, \tilde{u}_t)$, and the RM belief $g_\phi$ conditions on the observation-action history. 
\ttt{RMSM} then concurrently optimizes two independent objectives: prediction of a belief over RM states by minimizing a negative-log-likelihood loss $L_\phi$ and maximization of the policy's expected return $J_\theta$. 

\begin{algorithm}[t]
    \SetAlgoLined
    Inputs: URM-MDP $\mathcal{K}$, policy $\pi_\theta$, RM belief RNN $g_\phi$; 
    
    \tcc{Collect on-policy data}
    \vspace{1mm}
    $\mathcal{D} = \{\}$ \;
    \vspace{1mm}
    \For{$t$ from $1$ to $T$}{
        Compute RM state belief $\tilde{u}_t = g_\phi(\cdot | s_0,a_0,\ldots,s_{t-1},a_{t-1}, s_t)$ \;
        Execute action $a_t \sim \pi(\cdot | s_t, \tilde{u}_t)$ and observe $r_t, s_{t+1}, u_{t+1}$ \;
        Add $(s_t, a_t, r_t, s_{t+1}, u_t, \tilde{u}_t)$ to buffer $\mathcal{D}$ \;
    }
    
    \tcc{Train RM belief RNN } 
    Update $\phi$ with SGD on $L_\phi = \mathbb{E}_{\tau\sim\pi_\theta, t\sim\mathcal{U}{ \{1 , \ldots, |\tau| \}}}[- \log {g_\phi(u_t | s_0, a_0, \ldots, s_{t-1},a_{t-1},s_t)}]$ estimated from $\mathcal{D}$ \;
    
    \tcc{Train policy}
    Update $\theta$ with SGA on $J_\theta = \mathbb{E}_{\tau \sim \pi_\theta}[\sum_{t=0}^\infty{\gamma^t r_t}]$ estimated from $\mathcal{D}$ \;
    \caption{Reward Machine State Modelling (for URM-MDPs; 1 iteration)}
    \label{algo:rmsm}
    
\end{algorithm}


\begin{subsection}{Theoretical Advantages}
\ttt{RMSM} offers a number of intuitive benefits over prior approaches. First, when the objective $\mathcal{L}_\phi$ is minimized with respect to $\phi$, $g_\phi$ returns the Bayes-optimal belief over RM states given the history $\pr(u_t | s_0, a_0, \ldots, s_{t-1}, a_{t-1}, s_t)$ for histories with positive probability under $\pi_\theta$. In contrast, recall that \ttt{Thresholding} and \ttt{Independent Belief Updating} may fail to predict the correct RM state belief in URM-POMDPs, even with the Bayes-optimal approximate labelling function $\hat{\mathcal{L^*}}$. Theorem~\ref{theorem:rmsm} in the Appendix further adds that under some assumptions, when both \ttt{RMSM} objectives are simultaneously optimized, the resultant policy is optimal in the URM-(PO)MDP.

\ttt{Recurrent Policy}, like \ttt{RMSM}, also optimizes an objective aligned with solving the URM-(PO)MDP. However, \ttt{RMSM} exploits the task structure to reduce the complexity of the RL problem, while \ttt{Recurrent Policy} treats the problem as a general POMDP. This is evident in the URM-MDP case, where \ttt{Recurrent Policy} trains a policy conditioned on the entire state-action history, while the \ttt{RMSM} policy only conditions on the current state $s_t$ and a compact RM state belief representation. In URM-POMDPs, both \ttt{RMSM} and \ttt{Recurrent Policy} must train policies conditioned on entire histories, but nonetheless, exposing the predicted RM belief to the policy may facilitate learning.


\end{subsection}

\begin{section}{Experiments}
\label{sec:experiments}

We conduct experiments to investigate the consequences of deploying RL policies without a perfect labelling function. We focus on three questions: 
\begin{enumerate*}[label*=\textbf{Q\arabic*:}, leftmargin=2em]
    \item Are prior methods that rely on approximations of the labelling function $\hat{\mathcal{L}}$ robust to errors in $\hat{\mathcal{L}}$?
    \item How does \ttt{RMSM} (ours) perform compared to prior approaches without a perfect labelling function?  
    \item How well do \ttt{RMSM} (ours), \ttt{Thresholding}, and \ttt{Independent Belief Updating} model the true RM state?
\end{enumerate*}

\def\drawgrid(#1)[#2]{
    \begin{scope}[shift={(#1)}, transform shape]
    

        \stone(3.77, 3.2)[80][0.9][1]
        \stone(3.77, 2.2)[80][0.9][1]
        \stone(3.77, 1.2)[80][0.9][1]
        \stone(3.77, 0.2)[80][0.9][1]
    
        \node[scale=0.7] at (0.75,3.2) {\faRobot};
        \node[scale=0.7] at (0.75,0.2) {\faHome};

        \foreach \i in {0,...,4} {
            \draw [black] (\i,0 - 0.05) -- (\i,4 + 0.05);  
        }
        \foreach \i in {0,...,4} {
            \draw [black] (0,\i) -- (4,\i);
        }
        \draw [black, line width=0.1cm] (0, 0) rectangle (4, 4) {};
        
        \foreach \i in {0,...,3} {
            \foreach \j in {0,...,3} {
                \pgfmathsetmacro{\myx}{#2[3 - \j][\i]}
                \node at (\i + 0.5, \j + 0.5) {\textbf{\myx}};
            }
        }

    \end{scope}
}

\def\errsu{{{0.21,0.15,0.12,0.87},{0.14,0.21,0.2,0.9},{0.26,0.23,0.15,0.84},{0,0.17,0.1,0.84}}}

\def\errsfp{{{0,0,0,1},{0.6,0,0,1},{0,0,0,1},{0,0,0,1}}}
        
\def\opacite{0.5}

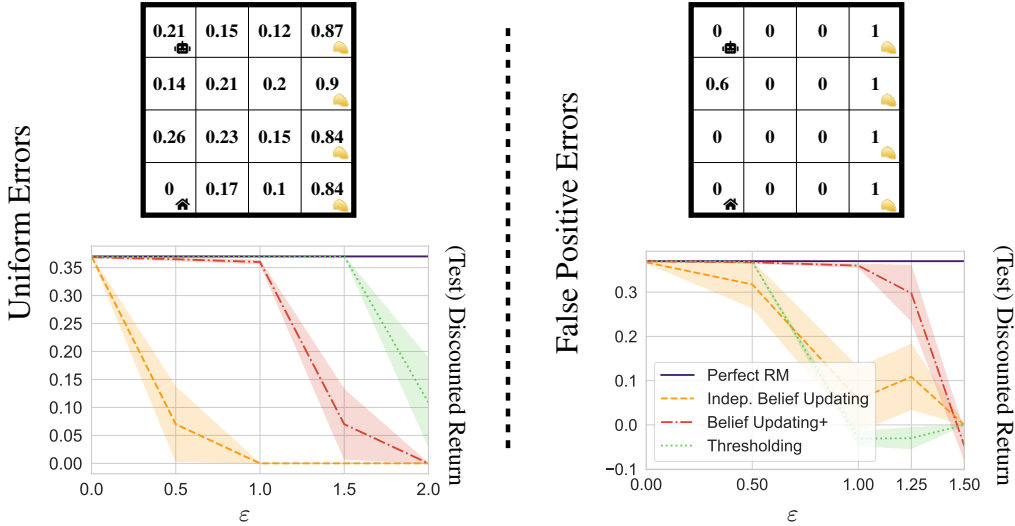
\begin{figure}[t]
    



    \vspace{-2cm}
    \begin{minipage}{.49\textwidth}
        \begin{tikzpicture}[
        myshade/.style={
            goldenrod!50,
            draw,
            ultra thin,
            rounded corners=.5mm,
            top color=gold!20,
            bottom color=gold!20!goldenrod
        },
        interface/.style={
            postaction={draw,decorate,
            decoration={border,angle=-135, amplitude=0.3cm,segment length=2mm}
            }
           }
        ]
        
        \scalebox{0.7}{
            \drawgrid(-0.5, 3)[\errsu]
        }
        
        \node at (1, 0){\includegraphics[width=.8\columnwidth]{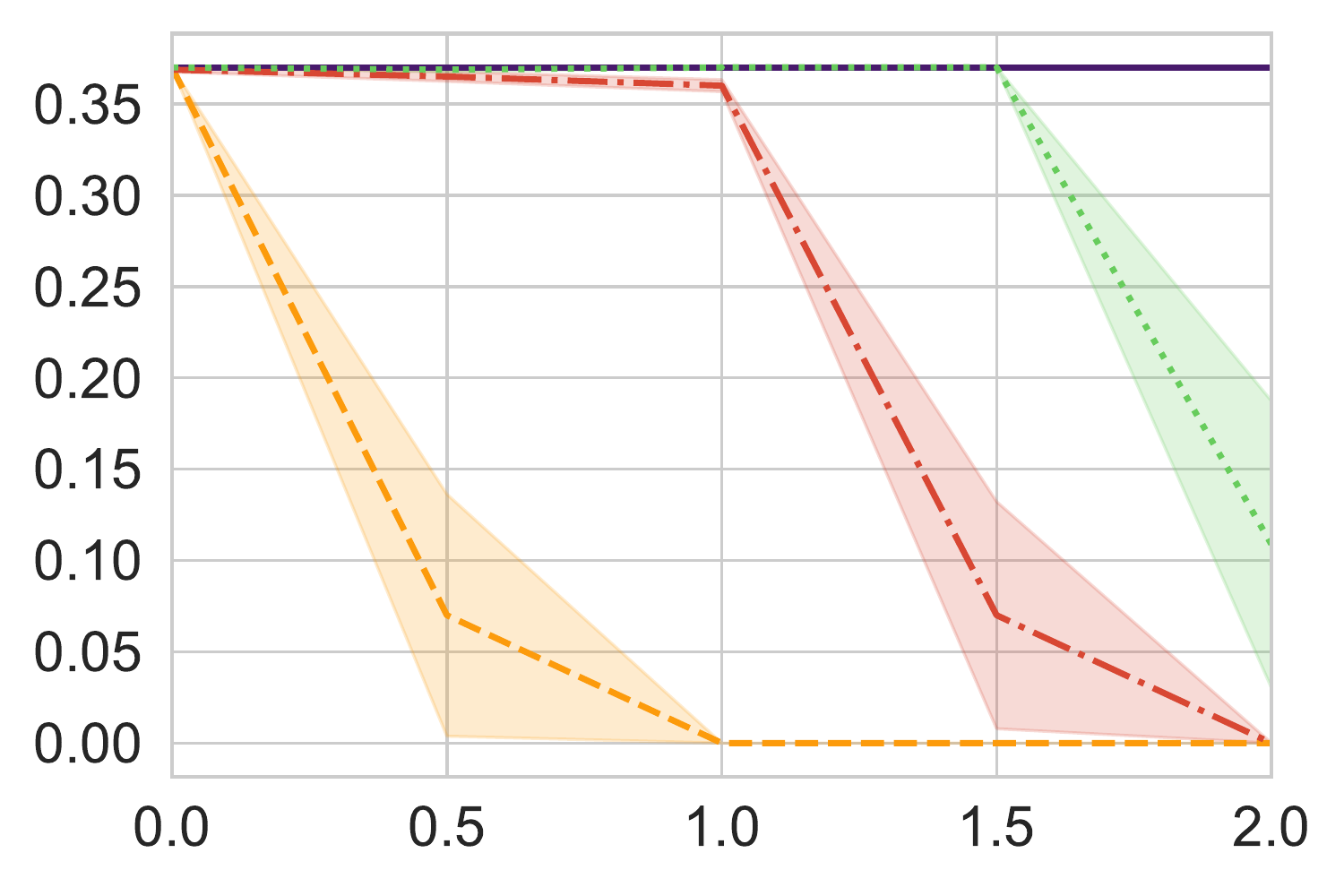}};
        \large
        \draw (-2,2) node[rotate=90] {Uniform Errors};
        
        \small
        \draw (1, -1.9)  node {$\varepsilon$};
        \draw (3.8,0.1) node[rotate=270] {(Test) Discounted Return};
        
        \draw [dashed, line width=0.06cm] (4.5,-1) -- (4.5,4.6);
        \end{tikzpicture}
    \end{minipage}
    \quad
    \begin{minipage}{.47\textwidth}
        \begin{tikzpicture}[
        myshade/.style={
            goldenrod!50,
            draw,
            ultra thin,
            rounded corners=.5mm,
            top color=gold!20,
            bottom color=gold!20!goldenrod
        },
        interface/.style={
                        postaction={draw,decorate,decoration={border,angle=-135,
                        amplitude=0.3cm,segment length=2mm}}
           }
        ]
        
        \scalebox{0.7}{
            \drawgrid(-0.5, 3)[\errsfp]
        }
        
        \node at (1, 0){\includegraphics[width=.8\columnwidth]{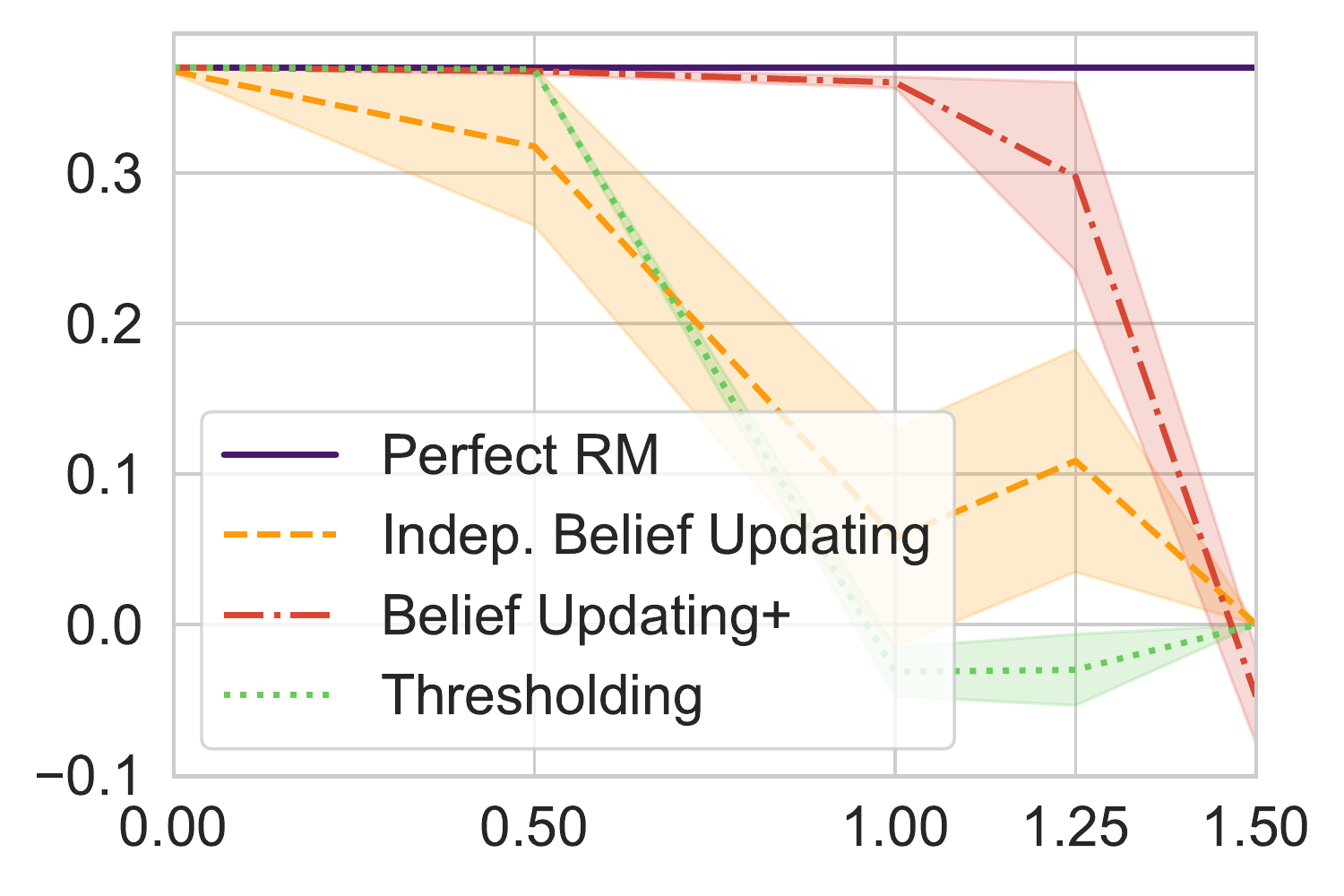}};
        \large
        \draw (-2,2) node[rotate=90] {False Positive Errors};
        
        \small
        \draw (1, -1.9)  node {$\varepsilon$};
        \draw (3.8,0.1) node[rotate=270] {(Test) Discounted Return};
        
        \end{tikzpicture}
    \end{minipage}
    
    \caption{The \emph{Mining} toy experiment. We consider two types of errors in predicting whether a square contains gold: \emph{uniform} (left), and \emph{false positive} (right). The parameter $\epsilon$ linearly controls the degree of error, with $\epsilon=0$ representing a ground-truth labelling function. Numbers in the grids represent the agent's probabilistic belief that a square yields gold for $\epsilon=1$. The graphs report mean final performances of various baselines along with standard error over 8 runs. 
    \ttt{Thresholding} is robust to small errors but fails when any square is misclassified, while \ttt{Independent Belief Updating} performs poorly as it ignores the dependence between propositions.}
    \label{fig:mining_result}
    \vspace{-1em}
\end{figure}

\begin{subsection}{Toy Gridworld Problem}

We investigate \textbf{Q1} on a toy grid domain with handcrafted models of $\hat{\mathcal{L}}$ and controllable levels of error. 

\textbf{Task:} \hspace{1mm} Returning to the \emph{Mining} example in Figure~\ref{fig:mining}, the agent must dig at a square containing gold and then deliver it to the depot to obtain a reward of 1, but visiting the depot without gold ends the episode. The agent does not observe when gold was obtained, but instead relies on a probabilistic belief $\hat{\mathcal{L}}(s, a_\mathrm{dig}, \cdot, \gold)$ that a state $s$ yields gold. We also introduce a cost of $-0.05$ for every movement action so that agents must carefully choose where to dig, instead of simply digging at every square.

\textbf{Approximate labelling function model:} \hspace{1mm} We consider two noisy models $\hat{\mathcal{L}}(s, a_\mathrm{dig}, \cdot, \gold)$  of whether each square contains gold shown in Figure~\ref{fig:mining_result}. The first gold model has a relatively uniform level of error for each square, while the second has a false positive belief that a single non-gold square may actually contain gold. For each, we control the degree of error through a parameter $\epsilon \in [0,\infty)$. $\epsilon=0$ corresponds to the ground-truth probabilities of gold (1 if the square contains gold and 0 if not), $\epsilon=1$ corresponds to the noisy probabilities reported in Figure~\ref{fig:mining_result}, and the probabilities are linearly interpolated for other values of $\epsilon$. $\hat{\mathcal{L}}$ is assumed to model \depot without error.

\textbf{Baselines:} \hspace{1mm} We consider \ttt{Thresholding} and \ttt{Independent Belief Updating} with each of the noisy gold models at various levels of error $\epsilon$. For comparison, we report the performance when using the ground-truth RM state (\ttt{Perfect RM}), and a belief updating approach supplied with the correct dependencies of the propositions over time (\ttt{Belief Updating+}). 

Precisely, \ttt{Belief Updating+} exploits the fact that the \gold proposition must maintain the same value each time you dig in state $s$. In other words, digging multiple times in the same state should not increase the agent's belief of having obtained gold. It is important to note that \ttt{Belief Updating+} cannot easily be applied beyond tabular domains.

Each method uses tabular Q-learning \citep{sutton2018reinforcement} to obtain a policy, with a linear approximation \citep{melo2007q} for methods conditioning on a belief over RM states. Hyperparameters can be found in the Appendix. 

\textbf{Results:} \hspace{1mm} Results are reported in Figure~\ref{fig:mining_result}. As expected, all methods performed near-optimally under a perfect labelling function ($\epsilon=0$), but both \ttt{Thresholding} and \ttt{Independent Belief Updating} showed sensitivity to errors in at least one case.
\ttt{Thresholding} performed well when errors were small in all states, but failed anytime an error exceeded the threshold, resulting in a misclassification of the proposition. 
\ttt{Independent Belief Updating} showed poor performance overall. We observed that this approach usually exploited errors in $\hat{\mathcal{L}}$ by repeatedly digging in the same square. However, \ttt{Belief Updating+} was significantly more robust to errors in $\hat{\mathcal{L}}$, showing the importance of considering the dependence between propositions over time.

\end{subsection}

\begin{subsection}{MiniGrid Problems}
\label{sec:minigrid_exps}

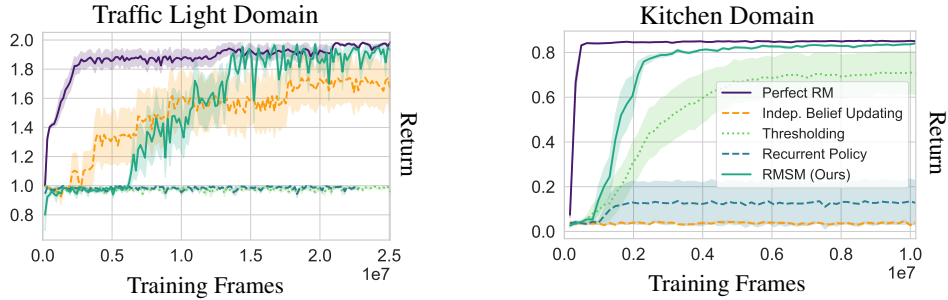
\begin{figure}[t]
    \centering
    \begin{minipage}{.49\textwidth}
        \begin{tikzpicture}
        \node at (0, 0){\includegraphics[width=.8\columnwidth]{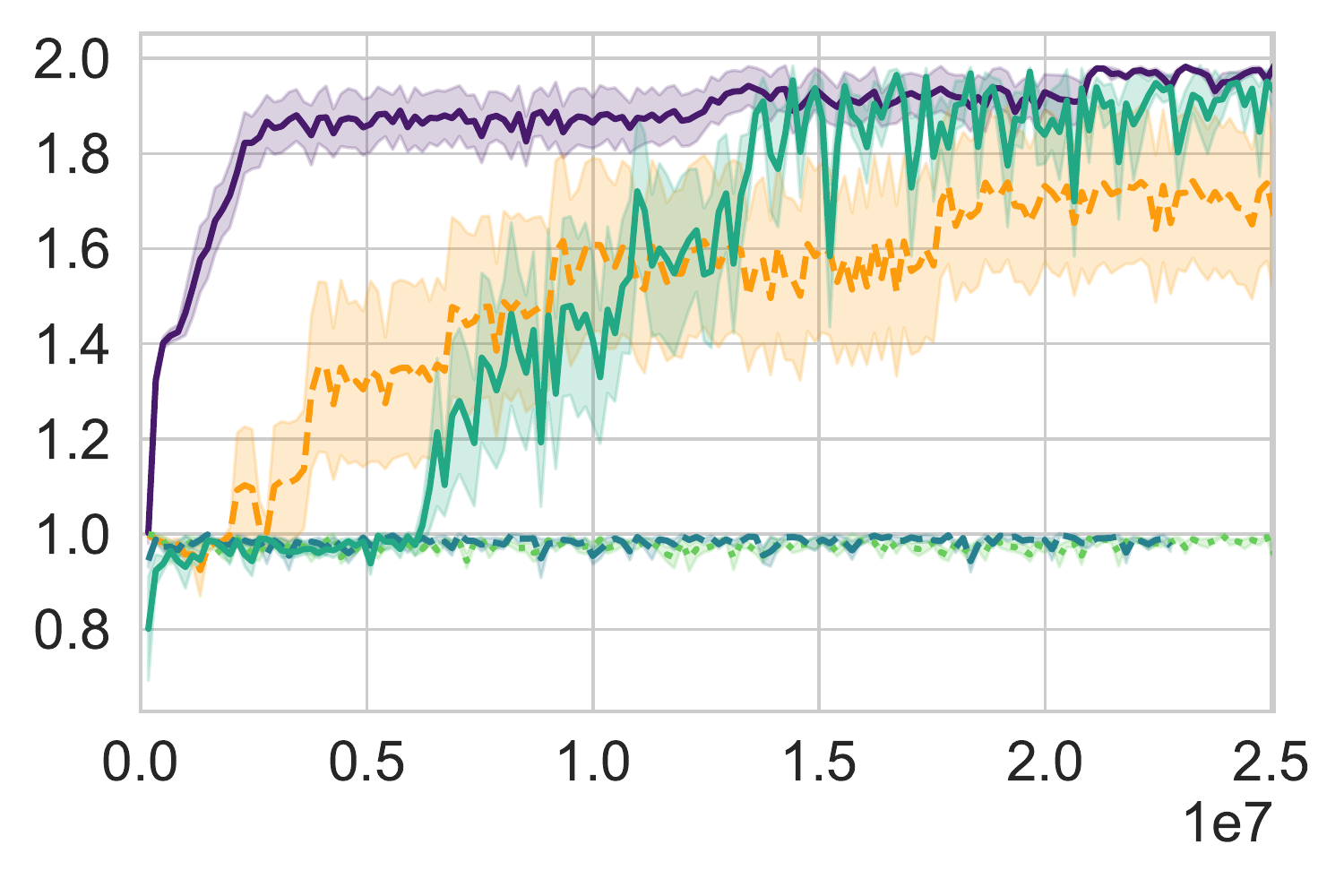}};
        \draw (0,1.9) node {Traffic Light Domain};
        \small
        \draw (0, -1.7)  node {Training Frames};
        \draw (2.7,0.2) node[rotate=270] {Return};
        
        \end{tikzpicture}
    \end{minipage}
    \begin{minipage}{.49\textwidth}
        \begin{tikzpicture}
        \node at (0, 0){\includegraphics[width=.8\columnwidth]{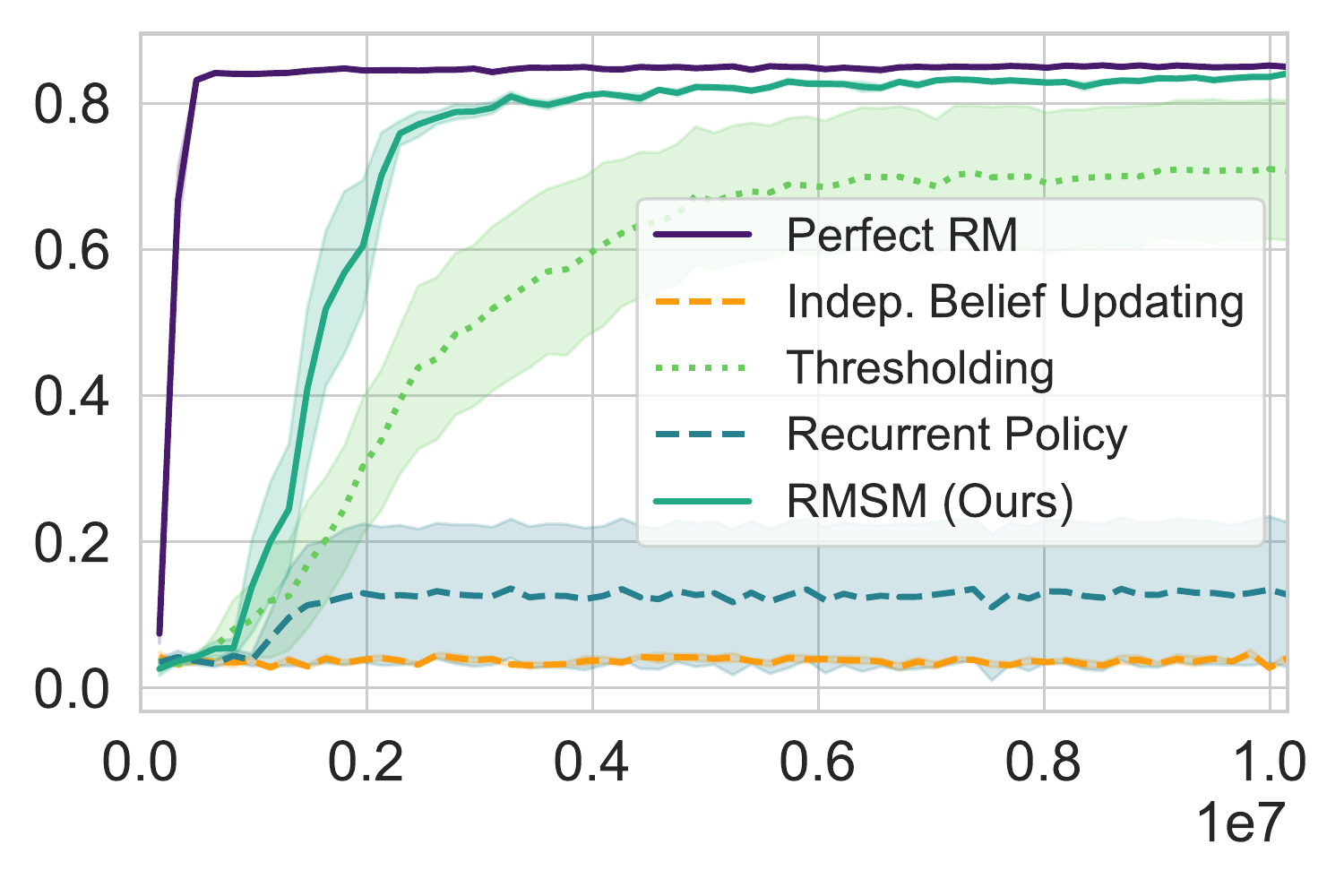}};
        \draw (0,1.9) node {Kitchen Domain};
        \small
        \draw (0, -1.7)  node {Training Frames};
        \draw (2.8,0.2) node[rotate=270] {Return};
        
        \end{tikzpicture}
    \end{minipage}
    
    \caption{Performance of \ttt{RMSM} (ours) vs various baselines on two partially observable, image-based domains (averaged over 8 runs, with 
    shaded areas showing standard error). \ttt{RMSM} converges similarly to when learning with the ground-truth RM state, while prior approaches fail on at least one domain.}
    \label{fig:results}
    \vspace{-1.6em}
\end{figure}

We investigate \textbf{Q2} and \textbf{Q3} on two partially observable, image-based MiniGrid environments encoding the \emph{Traffic Light} and \emph{Kitchen} problems from Section~\ref{sec:remarks-urm-pomdp}. Experimental details, including the RM specifications, network architectures, and hyperparameters can be found in the Appendix. 

\textbf{Baselines:} \hspace{1mm} We consider \ttt{RMSM}, \ttt{Thresholding}, \ttt{Independent Belief Updating}, \ttt{Recurrent Policy}, and an RL policy conditioned on the true RM state (\ttt{Perfect RM}). Policies are deep neural networks trained using PPO. The RM belief predictor $g_\phi$ (in \ttt{RMSM}) and the approximate labelling function $\hat{\mathcal{L}}$ (in \ttt{Thresholding} and \ttt{Independent Belief Updating}) are learned concurrently with the policy. 

\textbf{Results:} \hspace{1mm} We report learning curves in Figure~\ref{fig:results}. In both domains, \ttt{RMSM} rapidly learned a strong policy on-par with the policy that exploits ground-truth RM states. While \ttt{Thresholding} and \ttt{Independent Belief Updating} each made progress in one domain, they completely failed to learn in \emph{Traffic Light} and \emph{Kitchen}, respectively, matching our expectations from Section~\ref{sec:remarks-urm-pomdp}. \ttt{Recurrent Policy}, which does not exploit the structure of the problem, performed poorly despite enjoying similar theoretical guarantees to \ttt{RMSM}.  

To answer \textbf{Q3}, we compared beliefs over RM states predicted by \ttt{RMSM}, \ttt{Thresholding}, and \ttt{Independent Belief Updating} against a handcrafted approximation of the ground-truth belief (described in the Appendix). The final networks from training ($g_\phi$ for \ttt{RMSM} and $\hat{\mathcal{L}}$ for \ttt{Thresholding}, \ttt{Independent Belief Updating}) were used, and for an equal comparison, we conditioned each method on trajectories sampled under a random policy. Results in Figure~\ref{fig:bars} (in the Appendix) show that RM state beliefs produced by \ttt{RMSM} are much closer to ground-truth than those produced by \ttt{Thresholding} or \ttt{Independent Belief Updating}.


\end{subsection}



\end{section}

\begin{section}{Related Work}

Many recent works have considered specifying tasks for deep RL using RMs or a formal language, such as LTL. The vast majority of these assume access to a perfect labelling function (e.g. \cite{vaezipoor2021ltl2action, jothimurugan2021compositional, leon2021nutshell, liu2022skill, leon2020systematic}).
A few notable works in this vein learn policies that do not rely on a labelling function. \cite{kuo2020encoding} handle LTL tasks using a recurrent policy (inspiring the \ttt{Recurrent Policy} baseline), while \cite{andreas2016sketches, oh2017zero} and \cite{jiang2019language} learn modular subpolicies that are reusable between tasks.


Prior works have considered uncertainty in the labelling function, though most of this research comes from the motion planning literature, rather than the deep RL literature. However, these works typically assume that the state space is partitioned into discrete regions, resulting in a tabular MDP.  \cite{ding2011ltl} assumes that propositions in each state occur probabilistically in an i.i.d. manner and propose an approach exploiting this independence. \cite{ghasemi2020task} and \cite{verginis2022joint} consider a setting where propositions are initially unknown but can be inferred through environment interactions. 
\cite{cai2021reinforcement} considers learning policies that maximize the probability of satisfying an LTL task with uncertainties in the labelling function and environment dynamics. \cite{sadigh2016safe} and \cite{jones2013distribution} propose temporal logics over stochastic or partial observable properties.

More broadly, many prior works have considered the use of RMs or formal languages under uncertainty in the rewards, states, or ground-truth RM structure. \cite{xu2020joint, xu2020active} and \cite{gaon2019reinforcement} assume a setting where rewards are known but the RM structure is not, and propose algorithms to learn the RM.   
\cite{corazza2022reinforcement} learns RMs with stochastic reward emissions while \cite{velasquez2021learning} learns RMs with stochastic rewards and transitions. \cite{tor-etal-neurips19} learns RMs in partially observable environments, but defines the labelling function over observations rather than states, and thus does not consider uncertainty over propositions. \cite{shah2020planning} expresses the uncertainty over a desired behaviour as a distribution of LTL formulas.

\end{section}

\begin{section}{Conclusion}
In this paper, we studied the problem of learning Reward Machine policies under an interpretation of the vocabulary that is uncertain. We targeted our investigation towards complex, deep RL settings, where such uncertainty may naturally arise from sensor noise, modelling errors, or partial observability in the absence of a perfect labelling function. We first presented novel formulations of this problem as URM-MDPs and URM-POMDPs. Through illustrative examples and vision-based deep RL experiments, we then revealed a propensity for existing, naive approaches to exploit errors in their own model, leading to task failure and unsafe behaviour. To address these shortcomings, we proposed an algorithm, \emph{Reward Machine State Modelling}, that is cognizant of its uncertainty over atomic propositions and that exploits the structure of a URM-MDP or URM-POMDP to model the task-relevant RM states, rather than individual propositions. Empirical results demonstrated the robust performance of our approach without access to the ground-truth labelling function.


While we considered the problem of \emph{symbol grounding} from the viewpoint of Reward Machines, our insights are further applicable to a diversity of language-inspired abstractions for RL. We leave these extensions to future work.


\end{section}

\newpage
\section*{Acknowledgements}
We gratefully acknowledge funding from the Natural Sciences and
Engineering Research Council of Canada (NSERC), the Canada CIFAR AI Chairs Program, and Microsoft Research. Resources used in preparing this research were provided, in part, by the Province of Ontario, the Government of Canada through CIFAR, and companies sponsoring the Vector Institute for Artificial Intelligence (\url{https://vectorinstitute.ai/partners}). Finally, we thank the Schwartz Reisman Institute for Technology and Society for providing a rich multi-disciplinary research environment.

\bibliography{JAIR.bib}
\bibliographystyle{plainnat}

\appendix

\newpage
\appendix
\appendixpage

\section{Additional Definitions, Theorems, and Proofs}

\begin{definition}[\emph{Uncertain-Proposition Reward Machine POMDP} \textbf{(URM-POMDP)}]
\label{urmpomdp}
Given a POMDP (without reward function) $\mathcal{M} = \langle {S},\Omega, {T},{A},{P}, {O}, \gamma, \mu \rangle$, a finite set of atomic propositions $\mathcal{AP}$, a (ground-truth) labelling function $\mathcal{L}: S\times A \times S \rightarrow 2^{\mathcal{AP}}$, and a reward machine $\mathcal{R} = \langle U, u_0, F, \delta_u, \delta_r \rangle$, a {URM-POMDP} is a POMDP $\langle S', \Omega, T', A, P', O', R', \gamma, \mu' \rangle $ with state space $S' = S \times U$, observation space $\Omega$, terminal states $T' = (T\times U)\cup (S\times F)$, action space $A$, transition probabilities $P'((s_{t+1}, u_{t+1}) | (s_t,u_t),a_t) = P(s_{t+1}|s_t,a_t) \cdot \mathbbm{1}[\delta_u(u_t, \mathcal{L}(s_t, a_t, s_{t+1})) = u_{t+1}]$, observation probabilities $O'(o_t|(s_t,u_t), a_t) = O(o_t|s_t,a_t)$, reward function $R'((s_t,u_t),a_t,(s_{t+1},u_{t+1})) = \delta_r(u_t, \mathcal{L}(s_t, a_t, s_{t+1}))$, discount factor $\gamma$, and initial state distribution $\mu'(s,u) = \mu(s)\mathbbm{1}[u = u_0]$.
\end{definition}

\bigskip

\begin{theorem}[Optimality of \ttt{RMSM}]
\label{theorem:rmsm}
Let $\mathcal{K}$ be an URM-MDP or URM-POMDP, $\pi_\theta$ a policy, and $g_\phi$ an RM state belief predictor as defined in the \ttt{RMSM} algorithm. Assume that every feasible trajectory in $\mathcal{K}$ occurs with non-zero probability under $\pi_\theta$. For neural networks $g_\phi, \pi_\theta$ of sufficient capacity, assume that $g_\phi$ is a global minimum of $L_\phi$ and $\pi_\theta$ is a global maximum of $J_\theta$. 
Then $\pi_\theta, g_\phi$ together form an optimal policy in $\mathcal{K}$, and $g_\phi$ predicts the ground-truth belief over $u_t$, given the history. 

\medskip
\emph{Proof.} First, consider when $\mathcal{M}$ is a URM-MDP. For any particular history $s_0, a_0, \ldots, s_{t-1}, a_{t-1}, s_t$ under $\pi_\theta$, 
$\mathbb{E}_{u_t \sim \pr(\cdot | s_0, a_0, \ldots, s_{t-1}, a_{t-1}, s_t)} [- \log {g_\phi(u_t | s_0, a_0, \ldots, s_{t-1},a_{t-1},s_t)}]$ is minimized when $g_\phi(u_t | s_0, a_0, \ldots, s_{t-1},a_{t-1},s_t) = \pr(u_t | s_0, a_0, \ldots, s_{t-1},a_{t-1},s_t)$ for all $u_t \in U$  by Gibbs' inequality. For $g_\phi$ with sufficient model capacity, minimizing $L_\phi = \mathbb{E}_{\pi_\theta}[- \log {g_\phi(u_t | s_0, a_0, \ldots, s_{t-1},a_{t-1},s_t)}]$ (with respect to $\phi$) requires that $g_\phi(u_t | s_0, a_0, \ldots, s_{t-1},a_{t-1},s_t) = \pr(u_t | s_0, a_0, \ldots, s_{t-1},a_{t-1},s_t)$ for all feasible histories $s_0, a_0, \ldots, s_{t-1}, a_{t-1}, s_t$, since all feasible histories have positive probability under $\pi_\theta$. In other words, $g_\phi$ matches the ground-truth belief over RM states given the history. Note that for URM-MDP's, the ground-truth belief over RM states assigns all probability mass to the true RM state at time $t$. 

\medskip
Now consider $\pi_\theta(a_t | s_t, \tilde{u}_t)$, where $\tilde{u}_t = g_\phi(\cdot | s_0, a_0, \ldots, s_{t-1}, a_{t-1}, s_t)$. As articulated above, $g_\phi$ assigns all probability mass to the ground-truth RM state $u_t^*$ at time $t$ and so the inputs to $\pi_\theta$ precisely mimic the transitions in the corresponding RM-MDP (where the true RM state is observable) to $\mathcal{K}$. Since $\pi_\theta$ is an optimal solution to $J_\theta$, it is an optimal policy in this RM-MDP, and hence must be optimal in $\mathcal{K}$ as well. 
 
\medskip
When $\mathcal{M}$ is a URM-POMDP, the proof that $g_\phi$ matches the ground-truth belief over RM states given the history remains virtually unchanged. On the policy side, the \ttt{RMSM} policy is of the form $\pi_\theta(a_t | h_t, \tilde{u}_t)$. Note, however, that there even exists an optimal policy in the URM-POMDP of the form $\pi^*(a_t | h_t)$. Regardless of $g_\phi$, maximizing $J_\theta$ (with respect to $\theta$) can always recover an optimal policy in $\mathcal{M}$ by ignoring $\tilde{u}_t$.
\qed

\end{theorem}

\section{Experimental Details}

\begin{subsection}{Toy Experiments}

\end{subsection}
In the toy \emph{Mining} experiments, all baselines were trained using tabular Q-learning. \ttt{Independent Belief Updating} and \ttt{Belief Updating+} also used a linear approximation to condition on continuous beliefs over RM states, where the probability of each RM state was a feature. All methods were trained for 1 million frames using learning rate $0.01$, discount factor $\gamma = 0.97$, and random-action probability $\epsilon = 0.2$. 

\begin{subsection}{Deep RL Experiments}

\begin{figure}[b]
    \centering
    \includegraphics[align=c,width=0.44\textwidth]{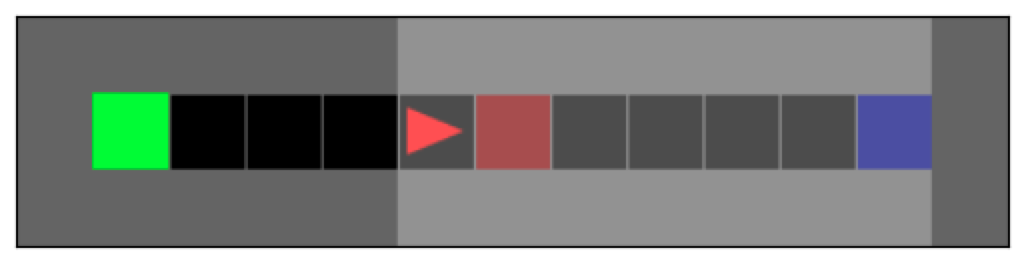}
    \qquad
    \includegraphics[align=c,width=0.36\textwidth]{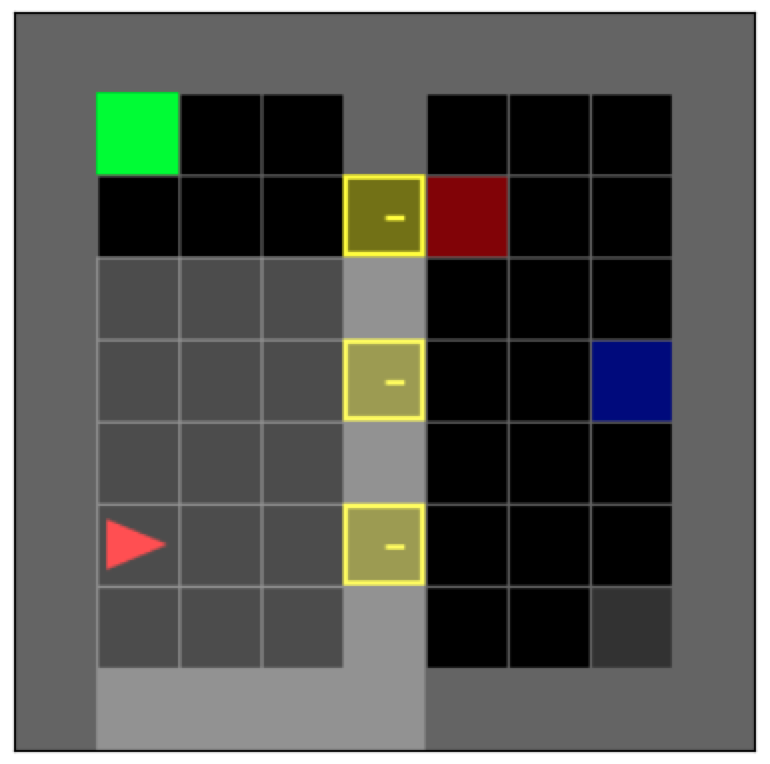}
    
    \caption{The \emph{Traffic Light} (left) and \emph{Kitchen} (right) tasks implemented in MiniGrid.}
    \label{fig:minigrid}
\end{figure}

\textbf{Task Descriptions:} \hspace{1mm} We encode the \emph{Traffic Light} and \emph{Kitchen}  tasks as MiniGrid environments, as shown in Figure~\ref{fig:minigrid}. In the \emph{Traffic Light} domain, the goal is to collect a package at the end of the road (represented by a blue square), and then return home (represented by the green square) without crossing a red light. The light colour stays green for a random duration, becomes yellow for one timestep, and then stays red for a random duration before returning to green. The agent can only observe squares ahead of itself, so the only safe way to cross the intersection is to check that the light is green before proceeding (at worst, the light becomes yellow while in the intersection, which is considered safe).   

We introduce propositions to determine if the agent is at the package, if the agent is at home, and if the agent is in the intersection during a red light. The RM state captures two pieces of information: whether the agent has crossed a red light at any time, and whether the agent has picked up the package. Reaching the package for the first time always yields a reward of 1. Returning home with the package ends the episode and yields another reward of 1. However, if the agent has crossed a red light, it instead receives a traffic ticket in the mail (and a -1 reward). 

In the \emph{Kitchen} domain, the agent must enter the kitchen, perform a set of chores (represented by the coloured squares in the right room, which turn gray when finished) by visiting them, before returning to the charging port at the top left. At the start, some chores may not need the agent's attention (e.g., if there are no dirty dishes) and that dish is considered done from the start. This occurs with a 1/3 probability for each chore, but the agent initially cannot observe which chores are done while outside the kitchen. 

Propositions for each chore determine on each timestep whether that chore is currently considered done, and an additional proposition determines if the agent is at the charging port. The RM for this task captures the subset of chores completed, resulting in $9$ possible RM states (including an additional terminal state for returning to the charging port). Returning to the charging port ends the episode, giving a reward of 1 if all chores are complete and 0 otherwise. We also introduce an energy cost of -0.05 for opening the kitchen door and performing each chore (including if it's already done, e.g. if the agent rewashes clean dishes) so that the agent must make careful decisions. 

\medskip
\textbf{Policy Architectures:} \hspace{1mm} \ttt{Perfect RM}, \ttt{Independent Belief Updating}, \ttt{Thresholding}, and \ttt{RMSM} each learned a policy conditioned on the current observation $o_t$ and a belief estimate of the RM state $\tilde{u}_t$ (or the ground-truth RM state, in the case of \ttt{Perfect RM}). We chose to condition policies on $o_t$ rather than the full history $o_0, a_0, \ldots, o_{t-1}, a_{t-1}, o_t$ to ensure that policies depended on the estimate $\tilde{u}_t$ to capture any relevant information from previous states.

Each policy consisted of the following components. Observations $o_t$ were encoded by a 16-channel convolutional layer, a max pooling layer, a 32-channel convolutional layer, and a 64-channel convolutional layer, in that order, each with kernel size $2 \times 2$ and stride $1$. Encoded observations and the belief over RM states $\tilde{u}_t$ were passed into an actor head with 3 hidden layers of $64$ units and ReLU activations, and a critic head with 2 hidden layers of 64 units and tanh activations.

\ttt{Recurrent PPO} used a similar architecture, except the history of observations were passed through an LSTM with hidden size 64.

\textbf{RM State Belief Prediction Architectures:} \hspace{1mm} \ttt{Independent Belief Updating} and \ttt{Thresholding} each learned an approximation of the labelling function $\hat{\mathcal{L}}$ by predicting $|\mathcal{AP}|$ logits (representing the probabilities of each binary proposition) conditioned on the history of observations. Observations were encoded by a similar architecture as in the policy, except the final output channel was 16. The sequence of observation encodings were passed into an LSTM with hidden size 16. Final predictions were decoded from the LSTM embedding using a single hidden layer of size 16. The predicted probabilities of propositions were used to derive estimates or beliefs over RM states, as outlined in Section~\ref{sec:methods}. 

\ttt{RMSM} used the same architecture as the approximate labelling function above, except the predicted outputs were $|U|$ logits, representing a Categorical distribution over $|U|$ reward machine states. 

\textbf{Hyperparameters:} \hspace{1mm} All policies were trained using PPO (\cite{schulman2017proximal}) for a total of 25 million frames in \emph{Traffic Light} and 10 million frames in \emph{Kitchen}. Hyperparameters are reported in Table~\ref{table:hyperparams}. For \ttt{Recurrent PPO}, the number of LSTM backpropagation steps was 4. 

\begin{table}[t]
\small
\centering
\caption{Hyperparameters for deep RL experiments}
\label{table:hyperparams}

\begin{tabular}{*{3}c}

\cmidrule[\heavyrulewidth]{1-3}
PPO Hyperparameters & \emph{Traffic Light} &
  \emph{Kitchen}
   \\ 

\cmidrule[\heavyrulewidth]{1-3}

\multicolumn{1}{l}{Env. steps per update} & 16384 & 16384  \\ 
\multicolumn{1}{l}{Number of epochs} & 8 & 8 \\ 
\multicolumn{1}{l}{Minibatch size} & 256 & 256 \\ 
\multicolumn{1}{l}{Discount factor ($\gamma$)} &  0.97 &  0.99   \\ 
\multicolumn{1}{l}{Learning rate} & $3\times10^{-4}$ & $3\times10^{-4}$ \\
\multicolumn{1}{l}{GAE-$\lambda$} & 0.95 & 0.95 \\
\multicolumn{1}{l}{Entropy coefficient} & 0.01 & 0.01 \\
\multicolumn{1}{l}{Value loss coefficient} & 0.5 & 0.5 \\
\multicolumn{1}{l}{Gradient Clipping} & 0.5 & 0.5  \\
\multicolumn{1}{l}{PPO Clipping ($\varepsilon$)} & 0.2 & 0.2 \\

\cmidrule[\heavyrulewidth]{1-3}
$\hat{\mathcal{L}}, g_\phi$ Hyperparameters (if applicable) & &
   \\ 

\cmidrule[\heavyrulewidth]{1-3}

\multicolumn{1}{l}{Env. steps per update} & 16384 & 16384  \\ 
\multicolumn{1}{l}{Number of epochs} & 16 & 8 \\ 
\multicolumn{1}{l}{Minibatch size} & 2048 & 256 \\ 
\multicolumn{1}{l}{Learning rate} & $3\times10^{-4}$ & $3\times10^{-4}$ \\
\multicolumn{1}{l}{LSTM Backpropagation Steps} & 4 & 4  \\

\bottomrule
\end{tabular}
\vspace{-5mm}
\end{table}

\textbf{Handcrafted RM State Belief:} \hspace{1mm} In Section~\ref{sec:minigrid_exps}, we compared the RM state beliefs of \ttt{RMSM}, \ttt{Independent Belief Updating}, and \ttt{Thresholding} against a handcrafted approximation of the true RM state belief. Note that the ground-truth RM state belief can be difficult to determine exactly.

Our approximation in \emph{Traffic Light} works as follows. Propositions for collecting the package and returning home can always be determined with certainty. When crossing the intersection, we sometimes know the light colour with certainty (e.g. if the agent observed the light was green on the previous timestep, it cannot be red this timestep). We consider all other crossings of the intersection "dangerous", where the light colour is uncertain. We approximate all "dangerous" crossings to have a $0.36$ probability of occurring during a red light, the steady-state probability of the light being red after sufficient time, absent any other knowledge. Using this model of the propositions, we determine an approximate belief of the current RM state given a history.

For \emph{Kitchen}, the agent has no information regarding whether each chore is initially complete while outside the kitchen. The belief over RM states during this period is easily determined using the known prior probabilities of each chore being complete. Once the agent enters the kitchen, we assume the RM state is known with certainty. Note that while this is often the case, the agent may sometimes enter the kitchen without looking at the state of all chores.
We again use this model of the propositions to derive an approximate belief of the current RM state given a history.

\begin{figure}[tb]
    \centering
    \begin{tikzpicture}
    \large
    \draw (-4.5,0) node[rotate=90] {Total variation distance};
    \node at (0, 0){\includegraphics[align=c,width=0.6\textwidth]{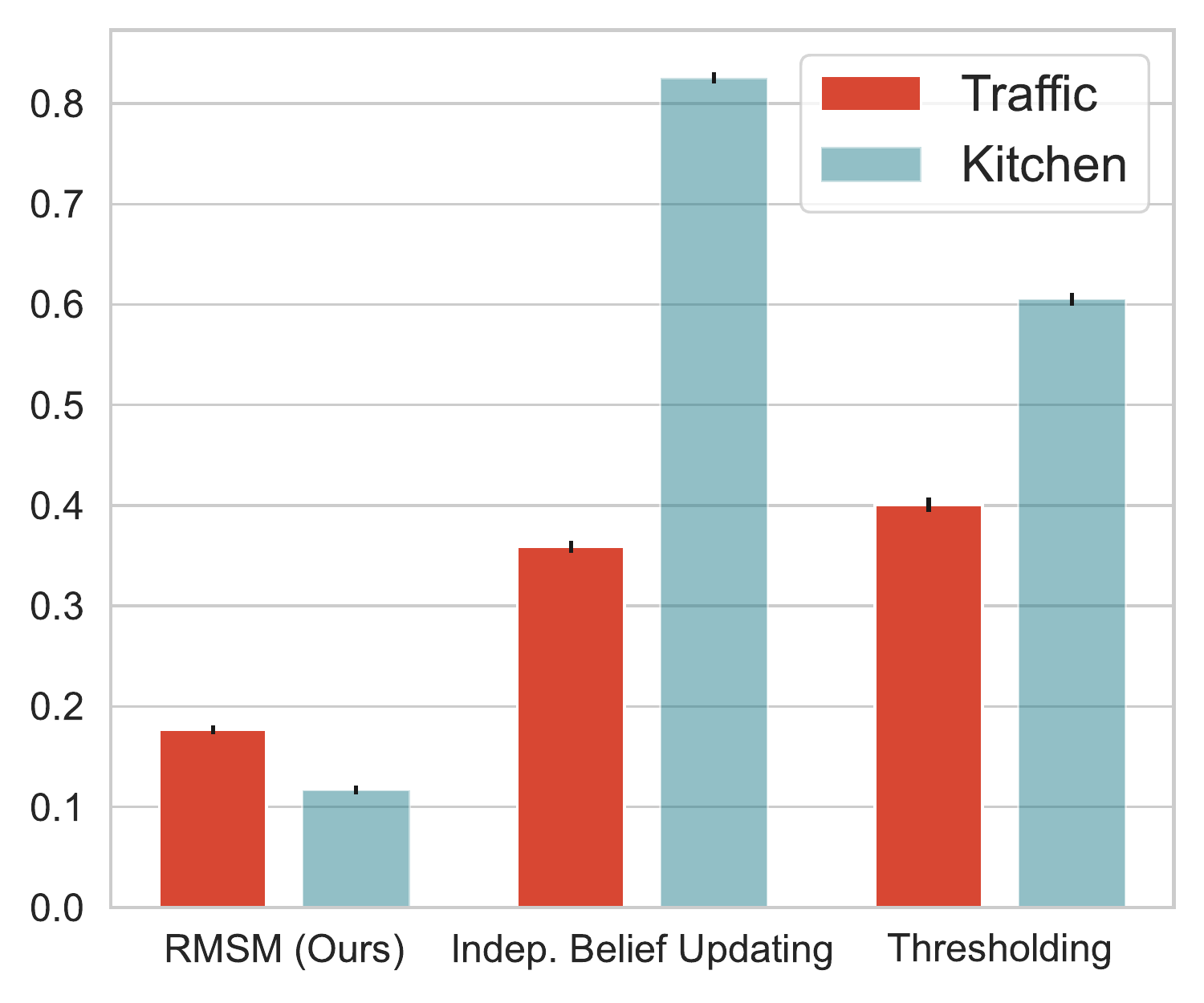}};
    \end{tikzpicture}

    \caption{The accuracy of RM state beliefs learned by \ttt{RMSM}, \ttt{Thresholding}, and \ttt{Independent Belief Updating} compared against a handcrafted approximation of the ground-truth RM state belief. Beliefs are conditioned on a history generated by a random policy. We report the Total Variation Distance between the predicted and (approximate) ground-truth beliefs, averaged over 8 seeds, 5 episodes per seed, and all possible histories in that episode. Error bars report standard error.}
    \label{fig:bars}
\end{figure}

\end{subsection}

\end{document}